\documentclass[man]{apa7}

\usepackage[american]{babel}
\usepackage{booktabs}
\usepackage{threeparttable}
\usepackage{csquotes}
\usepackage[style=apa,sortcites=true,sorting=nyt,backend=biber]{biblatex}
\usepackage{amsmath}
\usepackage{siunitx}

\newcommand{\NA}{\multicolumn{1}{c}{--}}

\DeclareLanguageMapping{american}{american-apa}
\title{Structure-Aware Modeling of Multiple-Choice Questions Improves Automatic Difficulty Estimation}
\shorttitle{MCQ structure and AQDE}

\author{Gabriel Ortega\textsuperscript{1}, Abelino Jiménez\textsuperscript{1}, Séverin Lions\textsuperscript{1, 2}, Pablo Dartnell\textsuperscript{1, 3, 4}}

\affiliation{\textsuperscript{1} Centro de Investigación Avanzada en Educación (CIAE), Instituto de Estudios Avanzados en Educación (IE), Universidad de Chile.\\
\textsuperscript{2} Departamento de Evaluación, Medición y Registro Educacional (DEMRE), Universidad de Chile.\\
\textsuperscript{3} Centro de Modelamiento Matemático (CMM), Universidad de Chile.\\
\textsuperscript{4} Departamento de Ingeniería Matemática (DIM), Universidad de Chile.
}

\abstract{
Automatic Question Difficulty Estimation (AQDE) holds growing promise for educational assessment because it has the potential to yield difficulty estimates that are competitive with expert judgment, while helping reduce the time and financial burden associated with pilot administrations and scaling to digital testing contexts. Prior AQDE studies report mixed evidence on whether adding distractors as additional text to the question stem and the correct key consistently improves difficulty prediction. We hypothesize that the effectiveness of distractor information depends on its structural representation, and that explicitly modeling distractors as separate components improves difficulty estimation over baselines that omit this information. To address this, we designed controlled architectures that model MCQ components as distinct inputs to isolate the contribution of distractor content and order. Specifically, we represented distractors by encoding each distractor as its own text input and aggregating their representations either with order-aware concatenation (with positional tags) or with an order-invariant summation. We evaluated these architectures using two Chilean datasets (Natural and Social Sciences, 2016--2020; 4{,}114 multiple-choice questions). Compared to a simpler model that only used the question stem and the key, our best distractor-aware architecture achieved higher predictive performance, reaching $R^2 = 0.83$ for Natural Sciences and $R^2 = 0.71$ for Social Sciences items. An order-invariant variant achieved nearly the same accuracy with approximately half as many parameters, offering a favorable accuracy-efficiency trade-off. These results show that structural information (especially distractor content) drives gains in predictive accuracy, supporting the development of efficient, structure-aware models that are computationally viable for large-scale educational applications.
}

\begin{document}
\maketitle

\section{Introduction}

As automatic systems for predicting item difficulty have become increasingly accurate and diverse, a central unresolved question is where their predictive power actually comes from. Specifically, does most predictive information reside in the stem alone, or do the answer options, and in particular the content of the distractors and their organization (e.g., the order in which they are presented), provide additional information that current models may only exploit implicitly? Question difficulty estimation refers to assigning either a numerical (continuous) or a categorical (e.g., easy/medium/hard) value to an item’s difficulty and is considered essential for data-driven test development. Within the framework of Item Response Theory (IRT), and specifically the Rasch model, precise difficulty parameters allow test developers to assemble forms with an appropriate distribution of item difficulties, ensuring adequate coverage of the examinee ability range and enabling the precise, orderable location of individuals on a measurement scale \parencite{de2013theory,van2005linear}. However, obtaining such estimates through large-scale field tests requires considerable time, personnel, and funding. Various approaches to Automatic Question Difficulty Estimation (AQDE) have been explored \parencite[see][]{benedetto2023survey,alkhuzaey2024text}, and a central question emerging from this literature is whether AQDE tools could partially or fully replace the piloting of multiple-choice questions (MCQs), thus accelerating calibration cycles and reducing associated costs \parencite{hsu2018automated}. When pilot data are lacking, developers often rely on domain experts' ratings; yet, even with rigorous training, such ratings remain slow, costly, and align only moderately with empirically estimated difficulty \parencite{sayin2024difference}. If reliable, AQDE would not only facilitate item calibration, but would also generate the real-time difficulty estimates required to administer optimally targeted items in computerized adaptive testing \parencite{van2000computerized}.

Given the growing interest in AQDE, an important question is how the internal structure of MCQs can be leveraged to improve prediction. In applied measurement, an MCQ is understood as a functional structure where the plausibility of distractors alters the discrimination demand among options \parencite{haladyna2002review,papenberg2017small}, systematically influencing item difficulty \parencite{ascalon2007distractor}. Because distractors act as functional components rather than mere supplementary text, it is reasonable to expect that explicitly modeling their content should yield relevant information for predicting Rasch difficulty. Aligning with this perspective, early conceptual work hypothesized that difficulty could be modulated by manipulating these structural components, especially semantic relationships between the stem, the key (correct answer) and the distractors. Ontology-based question generation approaches were among the first to formalize this idea: they articulated accessibility (stem–key similarity) and confusion (key–distractor similarity) as complementary mechanisms likely to shape difficulty \parencite{alsubait2014generating}. As text-based AQDE emerged, early models focused primarily on the stem. They combined handcrafted linguistic indicators (e.g., readability and cohesion) with coarse item-format descriptors (e.g., MCQs vs.\ true/false) to explain variance in difficulty \parencite{el2017predicting}. With the introduction of machine learning methods, researchers began to incorporate the full set of answer options, typically as an undifferentiated block concatenated to the stem. This enabled bag-of-words representations (e.g., Term frequency – Inverse document frequency) to feed regressors such as Random Forests, but without encoding the distinct roles of the key and the distractors \parencite{benedetto2020r2de}. In parallel, more sophisticated approaches explicitly modeled the MCQ structure, separating stem–key and key–distractor relationships as distinct features \parencite{hsu2018automated}. Neural architectures such as dual-module networks then decomposed difficulty into recall (assessed by similarity to external knowledge) and confusion (captured through attention mechanisms that quantify pairwise interference between the correct option and the distractors), thereby explicitly modeling interactions among answer options \parencite{qiu2019question}. With the advent of deep learning, Transformer-based models were introduced into AQDE \parencite{zhou2020multi,qiu2019question,reyes2023multiple,benedetto2021application}. A common strategy in AQDE is to enrich the input by appending the correct option and/or all options to the stem. However, results are mixed: multiple studies report limited or inconsistent gains when adding distractors or full option sets \parencite{benedetto2021application,feng2025medical}, suggesting that simply increasing textual content does not necessarily translate into additional predictive information. Several AQDE studies explicitly vary input configurations (e.g., stem only, stem+key, stem+all options) and often report small or inconsistent gains from including distractors/options \parencite{benedetto2021application,feng2025medical}. These results indicate that ‘more text’ is not a sufficient condition for better difficulty prediction, motivating approaches that model option roles and structure more explicitly.

One plausible explanation is that unstructured concatenation does not force the model to represent the functional roles of options (key vs. distractors) nor the option–stem relationships that define plausibility. In such settings, distractor text may behave as noise (especially when distractors vary in plausibility or quality), masking any potential contribution. We address this identifiability limitation in unstructured option inclusion by enforcing role-aware representations and controlled comparisons. This design also makes it possible to separate two aspects of distractor representation that are often confounded: their textual content and their order. Because response-option placement can introduce incidental variation in MCQ responses \parencite{lions2023position}, distractor order is worth testing as a secondary modeling dimension. This is also consistent with recent evidence that option order can affect model behavior in multiple-choice settings \parencite{pezeshkpour2024large}. We evaluate whether preserving distractor order contributes useful predictive information once distractor content is represented explicitly. We therefore ask: under controlled architectures, what is the incremental value of modeling distractors as a distinct structural component, separately from the stem and key, and does representing distractors as an order-invariant set versus an ordered sequence matter?

Previous research has emphasized novel neural architectures and top-line metrics rather than isolating how specific MCQ components drive prediction accuracy \parencite{alkhuzaey2024text,benedetto2023survey,peters2025text}. Because studies deploy heterogeneous text representation models, preprocessing pipelines, and evaluation protocols, even on the same benchmark datasets, direct performance comparisons are difficult, and conjectures about the importance of distractor order remain untested. In this study, to decouple model capacity from MCQ structural information, we employ a streamlined baseline that pairs embeddings with a single bidirectional Long Short-Term Memory neural network and a dense output layer, a widely used standard configuration for text processing in settings with non-massive, supervised datasets \parencite{huang2015bidirectional}. By varying the structural inputs supplied to this baseline, we test whether AQDE improvements are driven primarily by explicit distractor encoding and whether order-invariant aggregation can retain accuracy while reducing parameter counts. This design isolates the contribution of MCQ structure from model complexity and supports an accuracy-efficiency analysis grounded in controlled model comparisons. We reconcile mixed prior findings on option inclusion by showing that gains emerge when distractors are modeled as a distinct component with explicit role information and controlled interaction pathways, rather than as undifferentiated appended text.
\section{Method}

\subsection{Data}
\begin{figure}[htbp]
    \centering
    \includegraphics[width=0.6\textwidth]{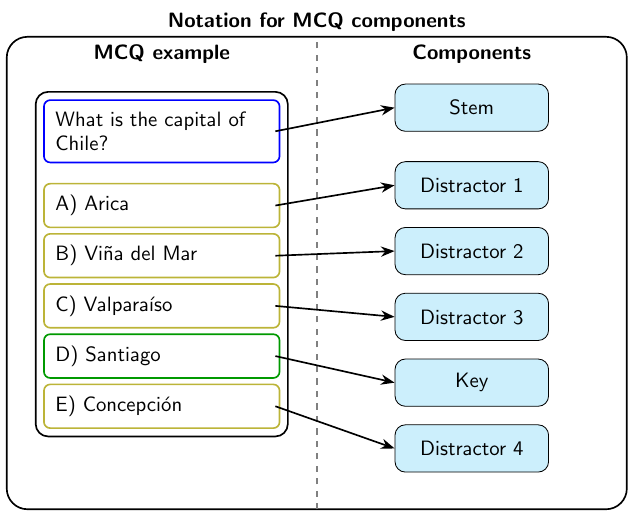}
    \caption{Components of a multiple-choice question (MCQ): stem, key, and distractors.}
    \label{fig:mcq-components}
\end{figure}

We used MCQs from Chilean national university-entrance exams as data for our AQDE experiments. Two datasets, consisting of the Natural and Social Sciences test forms administered from 2016 to 2020 by Chile’s Department of Evaluation, Measurement and Educational Records (DEMRE), were used (see Figure~\ref{fig:mcq-components} for an illustration of the anatomy of these MCQs). We used both datasets because they span distinct content domains and textual profiles, allowing us to test the generalizability of our approach across subject areas. Natural and Social Sciences examinations were delivered annually across the country, with approximately 150{,}000–200{,}000 candidates sitting the assessments each year. Because the Natural Sciences track allowed students to choose among Physics, Chemistry, Biology, or vocational–technical subtests, it yielded a greater variety of test forms and, consequently, more items than the Social Sciences track.

In total, the corpus contained 4{,}114 MCQs, 3{,}128 from Natural Sciences and 986 from Social Sciences, each labeled with a Rasch difficulty parameter estimated by DEMRE thanks to psychometric analyses of responses to pilot and official tests. The item difficulty parameters had a mean of 0.550 ($SD = 0.67$) for the Natural Sciences and 0.402 ($SD = 0.68$) for the Social Sciences, indicating that the Natural Sciences items were more difficult on average than the Social Sciences items. The disciplines also differed textually: the stems were more concise in Natural Sciences ($M = 39.1$ words) than in Social Sciences ($M = 83.6$ words). Similarly, the answer options in Natural Sciences were shorter ($M = 6.04$ words) than in Social Sciences ($M = 8.1$ words).

\subsection{Models}

\begin{figure}[htbp]
    \centering
    \includegraphics[width=0.9\textwidth]{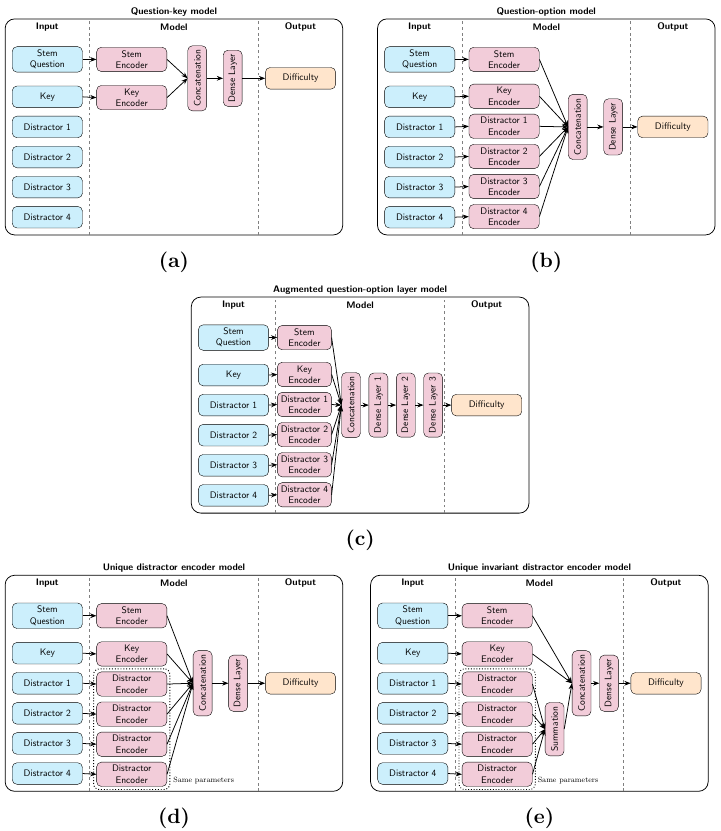}
    \caption{Five neural model variants for AQDE of MCQs. The models differ in how they encode and integrate the stem, key, and distractors: (a) Question-key model, (b) Question-option model, (c) Augmented Question-option model, (d) Unique Distractor Encoder model, and (e) Unique Invariant Distractor Encoder model.}
    \label{fig:architectures}
    \begin{flushleft}\footnotesize
    \textit{Note}. \textbf{(a)} Baseline without distractors (stem+key only). \textbf{(b)} Adds option-specific distractor encoders (no parameter sharing). \textbf{(c)} Increases post-concatenation interaction depth with extra dense layers. \textbf{(d)} Introduces a \emph{shared} distractor encoder with distractor order preserved via their relative position in the concatenation (parameter sharing as regularizer). \textbf{(e)} Uses the shared distractor encoder with order-\emph{invariant} summation across distractors.
    \end{flushleft}
\end{figure}

The experimental design evaluated five increasingly sophisticated model variants to study how the structural components of MCQs affect difficulty prediction (see Figure~\ref{fig:architectures}). All models followed the same high-level pipeline: (i) each available text field was mapped to a fixed-length vector by a text encoder; (ii) these vectors were combined into a single joint representation (typically by concatenation, or by an order-invariant aggregation); and (iii) a lightweight feed-forward prediction head transformed the joint representation into a scalar Rasch difficulty estimate. Within this common pipeline, the models differed only in how distractors were represented and combined, and in how much interaction capacity was added after combination. Concretely, we varied four factors: (1) whether distractors were included as separate inputs (vs.\ omitted), (2) how distractor representations were produced (independent encoders vs.\ a shared encoder with weight sharing across distractors), (3) how distractors were aggregated (order-aware concatenation vs.\ order-invariant summation), and (4) the depth of the post-aggregation prediction head (number of dense layers applied after the joint representation was formed). Accordingly, the Question-key model used only stem and key; the Question-option model added distractors as separate inputs with independent encoders (no weight sharing); the Augmented Question-option model increased the depth of the post-aggregation prediction head; the Unique Distractor Encoder model introduced weight sharing across distractors while preserving order via concatenation; and the Unique Invariant Distractor Encoder model retained weight sharing while replacing concatenation with an order-invariant summation over distractors. Unlike approaches that append all option text into a single sequence, we treat stem, key, and distractors as distinct inputs with functional roles. This design makes the contribution of each component identifiable under controlled comparisons, and helps interpret performance differences as changes in structural information rather than incidental properties of longer concatenated text.

Across all model variants, the text encoder adopted a modular design with four key components: word embeddings, batch normalization, bidirectional Long Short-Term Memory (BiLSTM), and dropout regularization. During model selection, we systematically compared a finite grid of encoder configurations, crossing batch normalization (enabled vs. disabled), dropout (0.5 vs. 0), and BiLSTM output (hidden state vs. full sequence), resulting in 12 encoder configurations for each model variant. Each encoder began with a 100-dimensional word-embedding layer with freely learned parameters, using words tokenized through a dictionary constructed from the training set vocabulary. The embedding outputs went through batch normalization to stabilize the activations and accelerate convergence. Next, a 64-unit per direction bidirectional LSTM network encoded contextual information from both directions of the sequence. A dropout rate of 0.5 could be applied to the LSTM output to mitigate over-fitting. The BiLSTM returned only the final hidden state or the full sequence of hidden states, producing a fixed-length vector representation of the entire input sequence. This systematic approach ensured that performance differences across our model variants were driven by the intended modeling choices, rather than by suboptimal hyperparameter settings.

The first model (Figure~\ref{fig:architectures}, panel a), the Question-key model, employed separate text encoder instances for the stem and the key, each with its own parameters, and concatenated their representations. This concatenated vector was first passed through a dense layer with 128 units and then to a final single-neuron output layer that produced the predicted Rasch difficulty parameter as a scalar value (linear activation in the output). This regression model directly treated difficulty as a continuous outcome, without any activation function in the output layer to allow for unconstrained prediction across the difficulty spectrum. The design purposefully omitted distractor information to establish a lower bound against which subsequent, more informative model variants could be judged. 

The next three models were designed as an interpretable progression, with each variant adding exactly one modeling dimension: (i) option-specific encoding, (ii) the depth of the prediction head after aggregation, and (iii) parameter sharing across distractors. First, the Question-option model (Figure~\ref{fig:architectures}, panel b) assigned an independent text encoder to every distractor, increasing the model's capacity to represent option-specific semantics before concatenating all representations, and passing them through the common prediction head, using the same prediction head as the Question–key baseline: a 128-unit dense layer followed by a linear output neuron. Second, the Augmented Question–option model (Figure~\ref{fig:architectures}, panel~c) retained the Question–option encoder configuration and used the same post-concatenation prediction head as the Question–key baseline, but then added two fully connected layers (128 and 64 units) before the final output, with the goal of increasing capacity to learn higher-order interactions among options (e.g., cross-option dependencies) after concatenation. Third, the Unique Distractor Encoder model (Figure~\ref{fig:architectures}, panel d) used a single text encoder for all distractors while preserving distractor order via their relative position in the concatenation, thereby testing the regularizing effect of parameter sharing, with the encoded representations concatenated and passed through the common prediction head (a 128-unit dense layer followed by a single linear output neuron). Together, these three models progressively extended the baseline by isolating different modeling dimensions, thereby setting the stage for the Unique Invariant Distractor Encoder model.

The fifth and most advanced model, the Unique Invariant Distractor Encoder model (Figure~\ref{fig:architectures}, panel e), retained the shared text encoder for distractors but replaced positional concatenation with a simple vector sum (element-wise addition) of the distractor encoder outputs. Specifically, we added the distractor output vectors to obtain a single pooled distractor representation. Because vector addition is commutative, this pooled representation was invariant to distractor ordering. The pooled distractor vector was then concatenated with the stem and key vectors and passed through the common prediction head, as in the other model variants.

\subsection{Training and Evaluation}

To ensure fair and controlled evaluation, we adopted a systematic protocol for data partitioning, model training, and performance assessment across all experiments. Each discipline-specific dataset (Natural Sciences and Social Sciences) was partitioned into training (56.25\%), validation (18.75\%), and test (25\%) sets \parencite{reyes2023multiple}. The validation set was used exclusively for early stopping of training based on validation performance, while the test set was reserved for the final performance assessment to prevent data leakage.

We applied the same training and optimization protocol to each model, using a single, fixed train–validation–test partition per discipline and reusing this partition across all training runs to ensure fair comparisons \parencite{benedetto2023survey}. For each model variant, we conducted an exhaustive search, exploring all 12 combinations of its optional components: batch normalization (enabled/disabled), dropout (enabled/disabled), and BiLSTM output (full sequence or final state). This yielded 60 distinct training runs and allowed us to evaluate each model variant under its optimal hyperparameter configuration.

Model performance was primarily assessed by minimizing the Mean Squared Error (MSE) between predicted and true Rasch difficulty values. Additionally, we computed the coefficient of determination ($R^2$) on the test set, in line with common practice in AQDE \parencite{benedetto2023survey,qiu2019question}. As a reference point, we included a simple constant predictor that output the discipline-specific mean difficulty estimated on the training set, providing a lower bound on performance when item text was ignored and matching common practice in AQDE evaluations \parencite{benedetto2023survey,benedetto2020r2de}. To assess computational efficiency, we also reported the number of trainable parameters for each model. 

Training was conducted using the Adam optimizer (learning rate 0.001), with MSE as the loss function. Models were trained with a batch size of 32 for up to 150 epochs, employing a validation-based early stopping criterion: we monitored validation loss and stopped training if it did not improve for 15 consecutive epochs, restoring the model weights from the epoch with the lowest validation loss \parencite{benedetto2021application,reyes2023multiple}.

For each of the five model variants, we selected the best-performing configuration based on validation set performance. Final results corresponded to these optimal models evaluated once on the unseen test set, following standard AQDE evaluation protocols: training and model selection using only the training and validation sets, and final performance (MSE and $R^2$) computed on the test set \parencite{benedetto2023survey}. This approach allowed us to isolate and analyze the impact of differences in MCQ structural information, rather than specific hyperparameter settings. This controlled design directly tested whether distractor information became beneficial when represented explicitly as a structural component, rather than implicitly embedded in a single concatenated input.
\section{Results}

To evaluate the impact of incorporating structural information into the models, we compared the $R^2$ values obtained by each model variant (Table~\ref{tab:results}). The simplest neural model, Question–key model, achieved $R^2$ values of .790 for Natural Sciences and .636 for Social Sciences, indicating a substantial improvement over the mean baseline. Adding explicit distractor encoding in the Question–option model further increased $R^2$ to .815 (Natural) and .693 (Social), showing that modeling each distractor individually was beneficial. However, introducing deeper interaction layers in the Augmented Question–option model reduced $R^2$ to .635 (Natural) and .555 (Social), indicating that the additional dense layers reduced performance and may reflect poorer generalization (e.g., potential overfitting) under our training regime. Isolating distractor content effects from distractor order in the Unique Distractor Encoder model yielded the highest $R^2$ values (.829 for Natural and .708 for Social), highlighting the benefit of distinguishing these two sources of information. Removing order information in the Unique Invariant Distractor Encoder model resulted in a slight decrease ($R^2 = .816$ for Natural and $R^2 = .699$ for Social), suggesting that while order contributed marginally, the main gains came from incorporating distractor-specific textual information. In summary, explicit distractor encoding and parameter sharing added predictive power, deeper post-concatenation interaction layers reduced test-set performance, and distractor order had a small but consistent effect.

\begin{table}[tbp]
\centering
\begin{threeparttable}
\caption{\textit{Predictive accuracy (MSE and $R^2$ on the test set) and trainable parameters by model}}
\label{tab:results}
\begin{tabular}{
l
S[table-format=1.3]
S[table-format=1.3]
S[table-format=2.2]
@{\hspace{14pt}}
S[table-format=1.3]
S[table-format=1.3]
S[table-format=2.2]
}
\toprule
& \multicolumn{3}{c}{Natural Sciences} & \multicolumn{3}{c}{Social Sciences} \\
\cmidrule(lr){2-4}\cmidrule(lr){5-7}
Model & {MSE} & {$R^2$} & {Params.\ (M)} & {MSE} & {$R^2$} & {Params.\ (M)} \\
\midrule
Mean baseline
& 0.450 & \NA & \NA
& 0.480 & \NA & \NA \\
\addlinespace[2pt]
Question--key
& 0.094 & 0.790 & 0.44
& 0.174 & 0.636 & 0.37 \\
Question--option
& 0.083 & 0.815 & 1.03
& 0.147 & 0.693 & 1.10 \\
Augmented Question--option
& 0.164 & 0.635 & 10.87
& 0.213 & 0.555 & 10.87 \\
Unique Distractor Encoder
& 0.081 & 0.829 & 1.03
& 0.140 & 0.708 & 1.03 \\
Unique Invariant Distractor Encoder
& 0.082 & 0.816 & 0.51
& 0.144 & 0.699 & 0.55 \\
\bottomrule
\end{tabular}
\begin{tablenotes}[para]\footnotesize
\textit{Note}. “Params. (M)” reports trainable parameters in millions. The Unique Distractor Encoder attains the highest $R^2$ in both domains, whereas the Augmented Question–option model, despite having $\approx 10$ M parameters, underperforms. The Unique Invariant Distractor Encoder model achieves similar accuracy with far fewer parameters ($\approx 0.5$ M).
\end{tablenotes}
\end{threeparttable}
\end{table}

In addition to predictive accuracy, we examined the number of trainable parameters required by each model (Table~\ref{tab:results}). The Question-key model was the most compact neural model, with 0.37M-0.44M parameters, while the Question-option model increased this to just over 1 million. The Augmented Question-option model was by far the largest, with nearly 11 million parameters, yet it did not yield better $R^2$ scores, highlighting that increased capacity did not translate into improved test-set performance in this setting. The Unique Distractor Encoder model achieved the highest $R^2$ values with about 1 million parameters. The Unique Invariant Distractor Encoder model further reduced the parameter count to around 0.51M-0.55M, with only a minor decrease in $R^2$, yielding the most favorable accuracy-efficiency trade-off among the models considered. Overall, these results show that incorporating distractor information can improve performance, but that more parameters do not necessarily yield better generalization.

Taken together, the results indicate that the highest predictive accuracy is obtained when distractor-specific textual information is incorporated. Including distractor order confers only a small gain in terms of $R^2$ ($\Delta R^2 \approx 0.01$-$0.013$ across domains) and at a higher parameter budget, while increasing model complexity with additional dense layers did not improve test-set performance in this setting. The most effective model variants are those that balance structural information with model simplicity, providing strong predictive power without unnecessary complexity.
\section{Discussion}

This study aimed to clarify how much the content and organization of the structural components of MCQs affect AQDE performance. We addressed the gap in the literature regarding the separate contributions of distractor content and distractor order, which are rarely isolated under controlled comparisons in the AQDE literature. By employing a controlled experimental design, where only the structural information provided to a baseline BiLSTM was systematically varied, we were able to isolate the impact of each component. This approach ensured that observed performance differences could be attributed to the structural features themselves, rather than to confounding factors such as model size or training protocol.

Our results demonstrated that the explicit modeling of distractors was the most important factor in our controlled setting to improve predictive accuracy in AQDE. Models that incorporated distractor information, especially when encoded as an order-invariant set, consistently outperformed the one that relied solely on the stem and key, as evidenced by substantial gains in $R^2$ across both Natural and Social Sciences datasets. In absolute terms, our best-performing models reached $R^2 = 0.83$ ($\text{Pearson } r \approx .91$) for Natural Sciences and $R^2 = 0.71$ ($\text{Pearson }r \approx .85$) for Social Sciences, values that lie at the upper end of those reported by recent text-based AQDE systems \parencite{peters2025text}. While cross-study comparisons are necessarily approximate due to differences in datasets, difficulty scales, and evaluation protocols, this pattern indicates that structure-aware models with explicitly modeled distractors can achieve predictive performance that is competitive with recent text-based AQDE systems without resorting to very large neural models. Notably, the best-performing model achieved these improvements with fewer parameters than more complex models, highlighting that structural modeling can enhance both computational efficiency and the fine-grained quantification of difficulty. These findings suggest that, for practical and scalable AQDE, representational alternatives for distractors can be as consequential as modest changes in the prediction head (e.g., adding post-concatenation dense layers), at least under controlled comparisons where the text encoder and training protocol are held constant. More broadly, the results indicate that leveraging MCQ structure explicitly can yield reliable gains without requiring very large models, though the relative importance of representation versus model capacity depends on the dataset and modeling regime.

While previous AQDE models have incorporated distractor information, our experiments quantify its contribution under controlled conditions in a way that isolates what is being added: distractor-specific textual representations, not additional model capacity or a different training regime. Across both domains, injecting distractor representations consistently improved predictive accuracy, which suggests that distractors carry systematic information about item difficulty beyond what is available in the stem and the key alone. From a measurement perspective, this is plausible because MCQs are functional structures in which distractors are not incidental text but competing options to the key \parencite{haladyna2002review,papenberg2017small}; when distractors are semantically plausible, they shape how readily examinees can identify the correct option and may therefore contribute information associated with Rasch difficulty \parencite{ascalon2007distractor}. In this sense, while not a direct validation of distractor quality, the performance gains obtained by explicitly encoding distractor text suggest the model becomes sensitive to the predictive information associated with distractor plausibility and functioning \parencite{haberman2019distractor,haladyna1993many}.

This aligns with research showing that leveraging distractor information yields better measurement. For instance, comprehensive distractor analysis using IRT-based models has shown that including distractor-level parameters leads to better fit and increased reliability in high-stakes assessments \parencite{haberman2019distractor}. From this perspective, the incremental accuracy obtained by encoding distractor text may be interpreted as reflecting information related to distractor plausibility (or attractiveness to examinees), although this property is not directly observed in our data. The central factor is the presence of functional distractors, those plausible enough to be attractive to a substantial number of examinees, which contribute meaningfully to an item's psychometric properties. Conversely, non-functioning distractors add little measurement value and can even degrade item quality \parencite{tarrant2009assessment,haladyna1993many}. Given that our data come from high-stakes national exams where distractors are professionally crafted, the success of our distractor-aware models provides large-scale, empirical support for this principle. In short, our findings suggest that well-crafted distractors, plausibly reflecting common student misconceptions, are important drivers of item difficulty, and that modeling them directly is associated with measurable gains in predictive accuracy.

Another important finding of this study is that distractor content is the primary driver of item difficulty, while its position plays a distinct, secondary role. This distinction is directly supported by our results: the order-invariant model (Unique Invariant Distractor Encoder model) achieved nearly the same predictive accuracy, with only a marginal drop in $R^2$, as its order-aware counterpart, while requiring only half the parameters. Importantly, this finding aligns with large-scale research on the same Chilean national test population, which reports positional effects in MCQs items that are systematic but small and that form a source of variation distinct from distractor attractiveness as studied in psychometric analyses \parencite{lions2023position}. Taken together, these empirical findings support a modeling strategy that treats distractor content and position as complementary but distinct sources of difficulty, enabling more accurate and efficient AQDE models.

Our findings may appear at odds with studies reporting limited gains from adding options/distractors via input concatenation. We interpret this difference as methodological rather than contradictory: when options are appended as undifferentiated text, the model is not forced to encode option roles (key vs. distractor) or the stem–option relations that define plausibility. In that setting, additional option text can behave as noise, especially if distractors are weak, redundant, or variable in quality, yielding negligible improvements. This likely arises from mechanisms such as role ambiguity, where concatenation blurs functional identities; insufficient relational pressure, where a single sequence does not explicitly encourage stem–option comparisons; and distractor quality variability, where non-plausible distractors increase text length without adding information. By explicitly modeling distractors as a separate component and contrasting order-invariant versus order-aware representations under a fixed protocol, we make the source of improvements more identifiable: performance changes can be attributed to structural information rather than to incidental properties of longer concatenated inputs. Practically, these results suggest that when item difficulty is driven by fine-grained option discrimination, AQDE models may benefit from role-aware representations that preserve option identity and encourage stem–option interaction. Conversely, when distractors are weak or not plausibly constructed, unstructured inclusion may add noise rather than information.

From a practical standpoint, our modeling approach offers significant advantages in feasibility and integration into test development workflows. A central advantage is computational efficiency: the best-performing models can be trained and run on standard personal computers. For example, the Unique Distractor Encoder, the most accurate variant, uses $1.03M$ parameters (approx.\ $4 \, MB$ of memory), while the more compact Unique Invariant Distractor Encoder achieves competitive results with only $0.51M$ parameters ($\approx 2 \, MB$). In contrast, typical Transformer-based models such as BERT-based models \parencite{zhou2020multi,benedetto2021application} contain approximately $110M$ parameters, often demanding specialized hardware (e.g., GPUs) and significantly greater computation time for both training and inference. This efficiency is particularly relevant for testing organizations where computational resources are limited, as it facilitates the adoption of automated item analysis tools without extensive infrastructure requirements. Furthermore, local processing supports the management of privacy-sensitive item content by reducing the need for external cloud-based services. 

Beyond feasibility and privacy, our models yield actionable outputs for practice, namely item-level predicted Rasch difficulties (and differences in predicted difficulty across model variants). Predicted difficulty can serve as an informative starting point for assembling forms that meet test blueprint targets (e.g., target difficulty distributions) and for narrowing the pool before piloting, complementing empirical calibration \parencite{van2005linear}. One downstream application of AQDE predictions is a simple diagnostic for distractor efficiency, leveraging differences between model variants rather than changing the AQDE training objective. Concretely, one can compare predictions from a model that ignores distractors with predictions from a model that includes them: a substantial positive shift in predicted difficulty when distractors are included would indicate that the distractor set, as a whole, is exerting meaningful influence, whereas a negligible or negative shift would suggest the presence of non-functioning distractors. We did not implement this diagnostic pipeline explicitly in the present study, but our comparison between model variants shows that adding distractor representations systematically changes predicted difficulty in ways that could be exploited for this purpose. This level of analysis could be integrated into routine item review cycles, helping to refine item quality and improve test fairness. In summary, our approach not only achieves high predictive accuracy but is also practical, offering actionable diagnostics that can be feasibly integrated into existing test development and validation processes.

Our study has limitations that open avenues for future research. First, our analysis of distractor-order effects is constrained by the nature of our dataset, which does not include items that differ only in the permutation of their distractors. While our findings suggest order is a comparatively minor factor in this setting, its true impact could be more precisely isolated with experimentally controlled data. Second, our items are structurally homogeneous (one key and four distractors), and our difficulty indices are derived from a single, consistent testing context and institution (DEMRE). Replicating these controlled comparisons on datasets from additional institutions (e.g., other testing agencies or item-development programs) and additional domains (especially Mathematics and Reading) is a necessary next step to assess cross-context generalization under comparable psychometric assumptions. Third, since many applied contexts report item difficulty using Classical Test Theory (CTT) indices, broadening the practical reach of AQDE beyond IRT-labeled settings will require linking or harmonizing external CTT-labeled item sources (e.g., proportion-correct indices from released item banks or large-scale assessments and classroom-oriented repositories) to an IRT-consistent scale using principled scale-linking strategies.

These limitations also clarify the psychometric scope of the present study. Our models predict empirically estimated Rasch difficulty, and the observed gains from distractor representations should be interpreted as evidence of predictive information rather than as direct validation of distractor functioning, item discrimination, or DIF. Extending the same structure-aware framework to DIF would therefore require a different empirical outcome. Given multi-group response data, one could compute item-level DIF statistics, for example, the Mantel--Haenszel common odds-ratio $\alpha$ or its associated $\Delta_{\text{MH}}$ index between focal and reference groups, and train the same modeling framework to approximate these indices from the stem, key, and distractors. In that setting, the same questions we ask here about the contribution of structural elements to difficulty (e.g., how much the stem, the key, and the ordering and plausibility of distractors matter) could be reformulated with a DIF outcome: rather than asking which structures make items harder or easier on average, we would ask which structures differentially advantage or disadvantage particular groups. Recent work shows that large language models can predict DIF from item text and, using explainable AI methods, highlight words that are most strongly associated with group differences in performance \parencite{maeda2025finding}. Our approach would complement this lexical perspective by emphasizing structural and option-level features of multiple-choice items. In future work, structure-aware AQDE could thus serve as a low-cost, early-warning screen to flag items that are not only unexpectedly difficult but also potentially inequitable, while high-stakes decisions remain grounded in established DIF methodology.

In conclusion, this study demonstrates that how a model encodes an MCQ's structure is a critical driver of AQDE accuracy. Our findings show that explicitly modeling distractors as unique semantic entities yields substantial performance gains, surpassing larger, more complex models with only a fraction of the number of parameters. This work shifts the focus from a pure ``bigger is better'' paradigm toward a more nuanced, structure-aware approach. By demonstrating that how a model represents a question can be as important as the model's overall complexity, we provide a clear design template for developing more accurate, efficient, and decomposition-oriented AQDE systems. The path forward lies not in building larger black boxes, but in designing models that intelligently leverage the inherent structure of the items they seek to understand.

\section{Data Availability Statement}
The data examined in this study cannot be shared publicly owing to confidentiality restrictions associated with proprietary test items from Chile's national university-entrance examinations administered by the Department of Evaluation, Measurement and Educational Records (DEMRE) at Universidad de Chile. 

\section{Funding}
This work was supported by ANID under the following grants: ANID CIAE CIA250005, ANID Fondecyt de Iniciación [11251579], FONDEF [IT25I0117] and ANID/PIA/Basal Funds for Centers of Excellence [FB210005] (Center for Mathematical Modeling).

\section{Acknowledgments}
The authors gratefully acknowledge the Department of Evaluation, Measurement and Educational Records (DEMRE) at Universidad de Chile for providing access to the item data used in this study.

\section{Conflicts of Interest}
The authors declare no conflicts of interest.
\printbibliography

\end{document}